\documentclass[letterpaper]{article} 
\usepackage{aaai23}  
\usepackage{times}  
\usepackage{helvet}  
\usepackage{courier}  
\usepackage[hyphens]{url}  
\usepackage{graphicx} 
\usepackage{natbib}  
\usepackage{caption} 
\usepackage{algorithm}
\usepackage{algorithmic}
\usepackage{amsmath}
\usepackage{amsthm}
\usepackage{amsfonts}
\usepackage{subcaption}
\usepackage{multirow}
\usepackage{booktabs}
\usepackage{arydshln}

\usepackage{newfloat}
\usepackage{listings}
\DeclareCaptionStyle{ruled}{labelfont=normalfont,labelsep=colon,strut=off} 
\floatstyle{ruled}
\newfloat{listing}{tb}{lst}{}
\floatname{listing}{Listing}
\title{GraphSR: A Data Augmentation Algorithm\\ for Imbalanced Node Classification}
\author{
    Mengting Zhou\textsuperscript{\rm 1,\rm 2}, Zhiguo Gong\textsuperscript{\rm 1,\rm 2}\thanks{Corresponding Author.}
}
\affiliations{
    \textsuperscript{\rm 1}State Key Laboratory of Internet of Things for Smart City, University of Macau, Macao\\
    \textsuperscript{\rm 2} Guangdong-Macau Joint Laboratory for Advanced and Intelligent Computing

    yb97402@um.edu.mo, fstzgg@um.edu.mo
}

\begin{document}

\maketitle

\begin{abstract}
	Graph neural networks (GNNs) have achieved great success in node classification tasks. However, existing GNNs naturally bias towards the majority classes with more labelled data and ignore those minority classes with relatively few labelled ones. The traditional techniques often resort over-sampling methods, but they may cause overfitting problem. More recently, some works propose to synthesize additional nodes for minority classes from the labelled nodes, however, there is no any guarantee if those generated nodes really stand for the corresponding minority classes.  In fact, improperly synthesized nodes may result in insufficient generalization of the algorithm.
	To resolve the problem, in this paper we seek to automatically augment the minority classes from the massive unlabelled nodes of the graph. 
	Specifically, we propose \textit{GraphSR}, a novel self-training strategy to augment the minority classes with significant diversity of unlabelled nodes, which is based on a Similarity-based selection module and a Reinforcement Learning(RL) selection module.
	The first module finds a subset of unlabelled nodes which are most similar to those labelled minority nodes, and the second one further determines the representative and reliable nodes from the subset via RL technique. 
	Furthermore, the RL-based module can adaptively determine the sampling scale according to current training data.
	This strategy is general and can be easily combined with different GNNs models.
	Our experiments demonstrate the proposed approach outperforms the state-of-the-art baselines on various class-imbalanced datasets.
\end{abstract}

\section{Introduction} 

Graph is regarded as one of the most powerful models for describing complex relationships between objects in various fields, such as natural language processing \cite{yao2019graph}, computer vision \cite{chen2019multi}, and recommendation systems \cite{wu2019session}. As the result, corresponding techniques for graph data analytics are receiving significant attention from the community. And GNNs (Graph Neural Networks) is one of the most successful techniques for node analysis. 
In principle, GCN \cite{kipf2016semi} aggregates node features in the spectral space using Laplacian matrices, while GraphSAGE \cite{hamilton2017inductive} aggregates features from node neighbors directly in the spatial domain. However, most existing GNNs are trained under the assumption that the node classes are balanced. Unfortunately, this assumption is not true in many real-world situations where some classes may have significantly few nodes than others in the training process. For example, the number of fraudsters in a social network is much smaller than that of benign ones in the fraud detection task \cite{liu2021pick}.
The class-imbalanced problem may cause the algorithm to bias towards the majority classes and ignore the minority classes in the representation learning. 
Therefore, it is challenging to apply GNNs directly to many real-world class-imbalanced graphs.

In fact, imbalance problem has been tackled for a long time and some significant progresses have been made in the area of machine learning. 
Generally speaking, the solutions can be summarized into three streams: data-level approaches, algorithm-level approaches, and hybrid approaches. 
Data-level approaches attempt to balance the class distribution by pre-processing training samples with over-sampling \cite{chawla2002smote} or under-sampling \cite{kubat1997addressing} techniques;
algorithm-level approaches take misclassification costs into consideration \cite{ling2008cost} or modify the loss function \cite{cui2019class} to alleviate the impact of class-imbalanced issue; and hybrid approaches combine above both \cite{batista2004study}. However, either data based or algorithm based method still solely relies on those labelled training data, therefore, inherently suffers from overfitting problem caused by the extensively reusing some samples or raising their weights. 


\begin{figure*}[htbp]
	\captionsetup[subfigure]{justification=centering}
	\centering
	\begin{subfigure}{0.28\textwidth}
		\includegraphics[width=\textwidth]{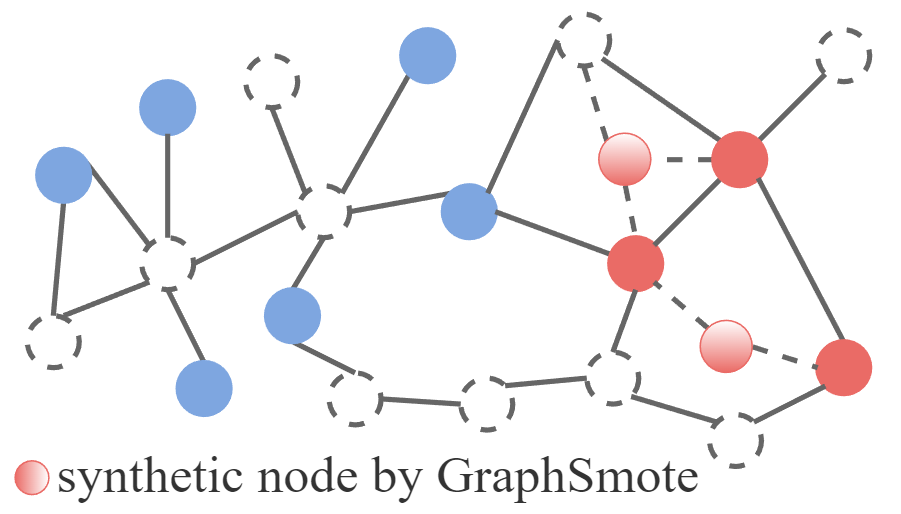}
		\caption{GraphSMOTE}
		\label{Fig.semi.1}
	\end{subfigure}
	\begin{subfigure}{0.28\textwidth}
		\includegraphics[width=\textwidth]{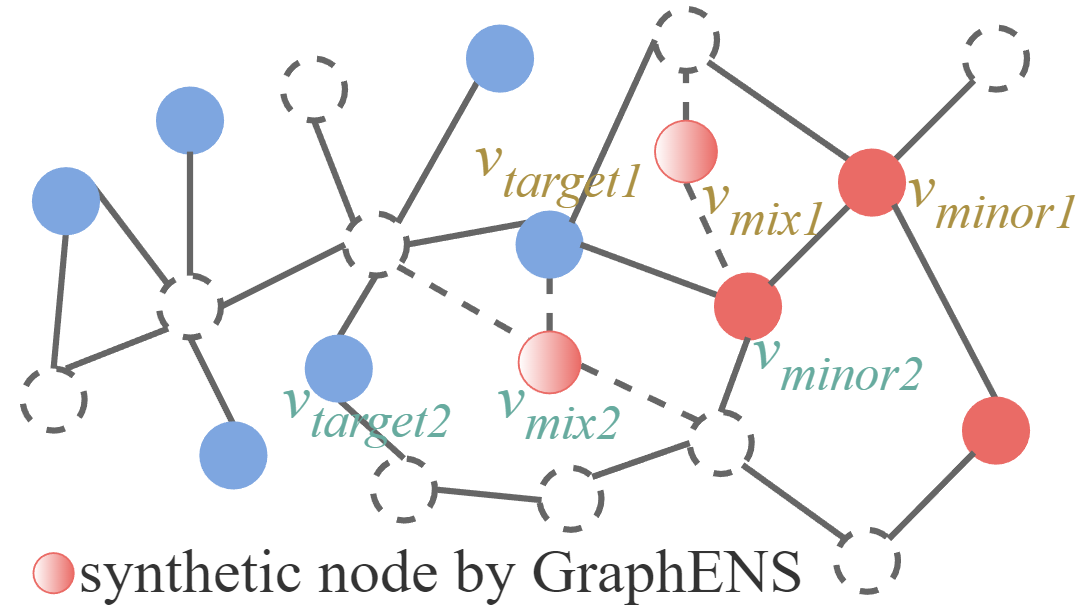}
		\caption{GraphENS}
		\label{Fig.semi.2}
	\end{subfigure}
    \begin{subfigure}{0.28\textwidth}
		\includegraphics[width=\textwidth]{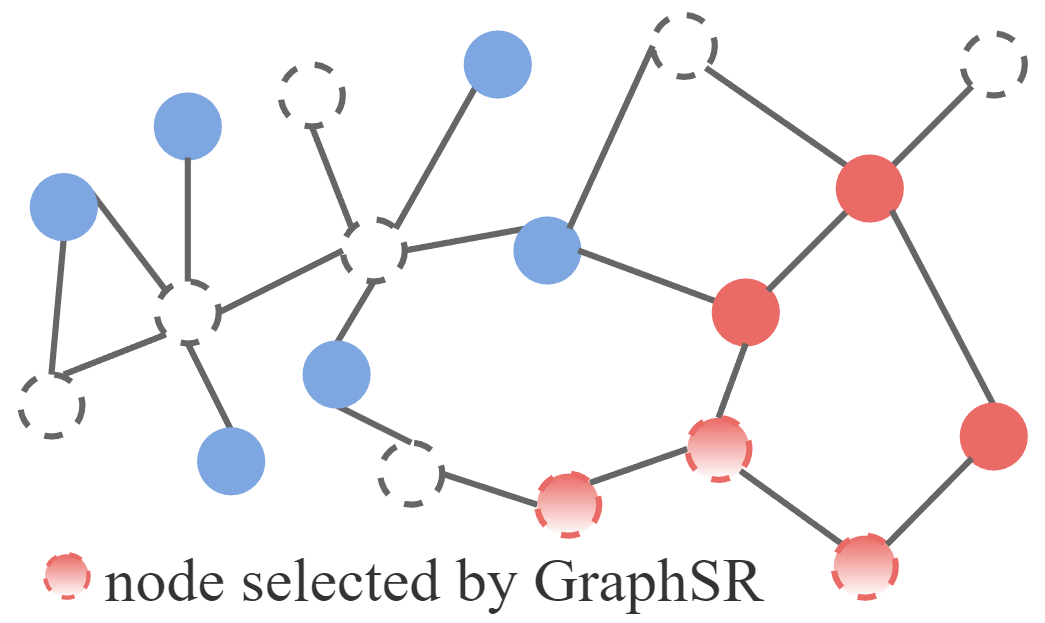}
		\caption{GraphSR}
		\label{Fig.semi.3}
	\end{subfigure}
	\caption{
		Here is a graph with limited labelled nodes and massive unlabelled nodes, blue nodes denote majority class, red nodes denote minority class and blank nodes are unlabelled nodes.
		(a) synthetic nodes are generated by two minority nodes and edges are generated by an edge predictor. (b) mixed nodes and their one-hop neighbors are generated by minor node and target node. (c) displays supplemental unlabelled nodes for minority class by GraphSR. }
    \label{fig1}
\end{figure*}

Actually, graph data naturally present topological structures of nodes, which can be used to generate some virtual nodes as data augmentation for the training algorithm. 
In this respect, recently GraphSMOTE \cite{zhao2021graphsmote} extend SMOTE \cite{chawla2002smote} to perform interpolation between two minority nodes in the embedding space to synthesize new samples, and exploit an edge predictor to determine the neighborhood of the synthetic samples as shown in Fig.\ref{Fig.semi.1}. Nevertheless, such generated nodes only rely on the minority nodes, which could not effectively extend the minority classes outwards (can still cause overfitting). 
To further deal with this, GraphENS \cite{park2021graphens} synthesize new minority nodes with their one-hop neighbors by mixing some existing minority node from other classes, thus, can enrich the diversity of those minority classes as shown in Fig.\ref{Fig.semi.2}. 
However, those synthetic nodes, which are generated under the subjectively designed mixing ratio between the minority nodes and other nodes, may not reveal the real situation of the underlying data nature, thus, could damage the results if the mixing ratio is not properly set.  
In addition, both of the above methods conduct over-sampling on minority classes heuristically with a fixed proportion, which could not be generalized across different datasets. The over-sampling ratio should be well-elaborated, otherwise the performance of majority classes will degrade when the ratio is set too large for minority classes.  

As a matter of fact, the previous works fail to utilize the rich information from the massive unlabelled nodes of the graph, which are valuable resources for generating more promising data to augment the minority classes. 
In the computer vision area, CReST \cite{wei2021crest} experimentally finds the phenomenon that minority classes suffer from low recall but achieve surprisingly high precision, thus minority pseudo-labels are less risky to supplement the training set. 
However, we empirically find that there are still many misclassified nodes in minority classes in the node classification task on graph, the details are shown in appendix.
Applying CReST directly on graph does not work well since the minority pseudo-labels are not reliable enough, and there is no mechanism to restrict these noisy unlabelled samples from being supplemented to the training set. Wrongly augmented data may damage the performance of the classification. 

Motivated by above discussions, in this paper we design a novel data augmentation algorithm for minority classes, called \textit{GraphSR}, as shown in Fig.\ref{Fig.semi.3}, where a reinforcement learning algorithm is employed to optimize the strategy for selecting those unlabelled data to augment the minority classes.  

Specifically, we firstly pre-train a baseline GNN model with labelled data, which can then generate pseudo-labels for those unlabelled data. 
Nevertheless, the baseline GNN model trained on imbalanced data could be biased towards majority classes, resulting in a poor prediction of unlabelled data.  
To tackle the problem, instead of supplementing the minority classes by randomly selecting some unlabelled nodes solely according to their  pseudo-labels, GraphSR firstly utilizes a similarity-based selection module to filter out the most similar unlabelled nodes for each minority class, which aims to efficiently discover potential nodes from the numerous unlabelled nodes and maintains a pool of potential data to augment those minority classes.
In the second step, in order to reduce the impact of noisy nodes, we design another module to adaptively choose the informative and reliable nodes from the candidate set through a reinforcement learning technique, named RL-based selection module.
In practice, GraphSR trains a selector as an agent to decide which node in the candidate set to be preserved, and then the action is evaluated by the environment with the improved classifier trained using the augmented dataset, the reward is assigned based on the performance of a class balanced validation set.
With the two-step selection, GraphSR can obtain the optimal unlabelled nodes to supplement the imbalanced training data. 
In this way, we can use the new training set to train an unbiased GNN classifier.

We summarize the main contributions as follows:
\begin{itemize}
    \item We propose to study the class-imbalanced node classification problem in the semi-supervised setting, where numerous unlabelled nodes can be exploited to supplement the minority classes.
    \item We design a novel data augmentation strategy, GraphSR, to efficiently sample the informative and reliable unlabelled nodes to enhance the diversity of the minority classes. The proposed method can adaptively determine the sampling scale based on the current training data, making it more generalizable to different datasets.
    \item Experimental results on several datasets show that the proposed approach outperforms all the baselines. More importantly, the technique can be injected into any of the GNNs algorithms.
\end{itemize}

\section{Related Work} 

\subsection{Class Imbalanced Learning}
Class imbalanced representation learning is a classical topic in machine learning domain and has been well-studied \cite{he2009learning}.
The goal is to train an unbiased classifier on a labelled dataset with a class-imbalanced distribution, where majority classes have significantly more samples and minority classes have fewer samples.
Prominent works include re-weighting and re-sampling approaches.
Re-weighting approaches try to modify the loss function by raising the weights of minority classes \cite{lin2017focal, cui2019class}, or expanding the margins on minority classes \cite{cao2019learning, liu2019fair, menon2020long}.
Re-sampling approaches attempt to balance the data distributions by pre-processing training samples deliberately, such as over-sampling minority classes \cite{chawla2002smote}, under-sampling majority classes \cite{kubat1997addressing}, and a combination of both \cite{batista2004study}. With the improvement of neural network, re-sampling strategies augment the minority classes through not only sampling techniques \cite{liu2020mesa}, but also generation idea \cite{kim2020m2m, wang2021rsg}.
The typical method SMOTE \cite{chawla2002smote} generates new samples by using interpolating technique over some minority samples and their nearest neighbors from the same class. While other works \cite{kim2020m2m, wang2021rsg} synthesize minority samples through transferring the common knowledge from majority classes.
However, most existing methods are devoted to i.i.d. data, and can not be directly utilized to graph-based data, where the relationships among objects should be considered.

\subsection{Graph Neural Networks}
Graph neural networks (GNNs) are firstly proposed in 2005 \cite{gori2005new}. With the rapid development of deep learning, the techniques have achieved enormous success in non-Euclidean structured data. 
Generally speaking, GNNs follow a message-passing scheme to recursively embed a node with its neighbors into a continuous and low-dimensional space \cite{gilmer2017neural}.
GNN techniques can be divided into two categories: spectral-based methods and spatial-based methods.
The spectral-based methods often apply the Laplacian matrix decomposition of the entire graph to collect nodes information ~\cite{defferrard2016convolutional,kipf2016semi,bianchi2020spectral}, while the spatial-based methods employ the topological structure of the graph directly and aggregate nodes features based on the topological information of the graph ~\cite{velivckovic2017graph,hamilton2017inductive,you2019position}.

For node classification task, there are some works proposed to deal with class-imbalanced issue.
DR-GCN \cite{shi2020multi} utilizes two types of regularization with class-conditioned adversarial training and latent distribution constraints on unlabelled nodes to train an unbiased classifier.
GraphSMOTE \cite{zhao2021graphsmote} extends SMOTE to the embedding space and combines it with edge generation to synthesize minority nodes.
ImGAGN \cite{qu2021imgagn} generates a set of synthetic minority nodes by modelling a generator to simulate both the attribute and topological distributions of the whole minority class.
PC-GNN \cite{liu2021pick} devises a label-balanced sampler to construct the sub-graphs, and chooses neighbors for each node in the sub-graphs by a neighborhood sampler for training.
GraphENS \cite{park2021graphens} synthesize the features and neighbors for minority nodes by mixing minority nodes and target nodes from other classes to avoid overfitting.
For the over-sampling works, GraphSMOTE and ImGAGN only adopt the identical minority nodes to synthesize new samples, it is prone to be overfitting, and the diversity is limited.
Although GraphENS utilizes the nodes from both minority classes and other classes when generating new samples, it changes graph structure and is hard to find out the optimal mixing ratio and neighbors, which may induce some noise and impact the performance instead.
Moreover, these methods over-sample the minority classes with a fixed ratio and fail to take full advantage of the abundant information available from unlabelled nodes.
In our work, GraphSR supplements the minority classes with unlabelled nodes to enrich the diversity and determines the sampling scale adaptively.

\section{Problem Definition} 
In this work, we target at semi-supervised class-imbalanced node classification on graphs, which is with a small ratio of labelled nodes and large amount of unlabelled ones.
We are going to use the limited number of labelled nodes for training a classifier, which is tested on the nodes from the same graph.
In our setting, each node belongs to only one class, and the distribution of classes is imbalanced, that is the majority classes have significantly more samples than those minority classes in the training set. 

Formally, an attributed graph is defined as $\mathcal{G}=(V,E,X)$, where $V$ is the nodes set, $E$ is the set of edges, and $X \in \mathbb{R}^{|V| \times d}$ denotes the attribute matrix where each row represents a $d$-dimensional attribute of the corresponding node. $\mathcal{N} (v)=\{v' \in V | \{v',v\} \in E\}$ is the set of the neighboring nodes that directly connect to $v$. Each node only belongs to one class $y$, and there are totally $m$ classes in the graph.
During training, only a subset of nodes $V_L$ with their corresponding labels $Y_L$ are available, and the unlabelled set of notes ia denoted as $U$.
For the labelled nodes, we denote their class distribution as $\{C_1, ..., C_m\}$, where $C_i$ is the node set of $i$-th class. Besides, we introduce an imbalance ratio $\rho $ to measure the degree of imbalance, $\rho = \frac{min_i(|C_i|)}{max_i(|C_i|)} $.

Given $\mathcal{G}$ with a labelled node set $V_L$, which is class imbalanced, we aim to train an unbiased classifier $f$ that can work well for the entire classes, with the help of unlabelled nodes in $U$.
 
\section{Methodology}
\begin{figure*}[t]
    \centering
    \includegraphics[width=0.95\textwidth]{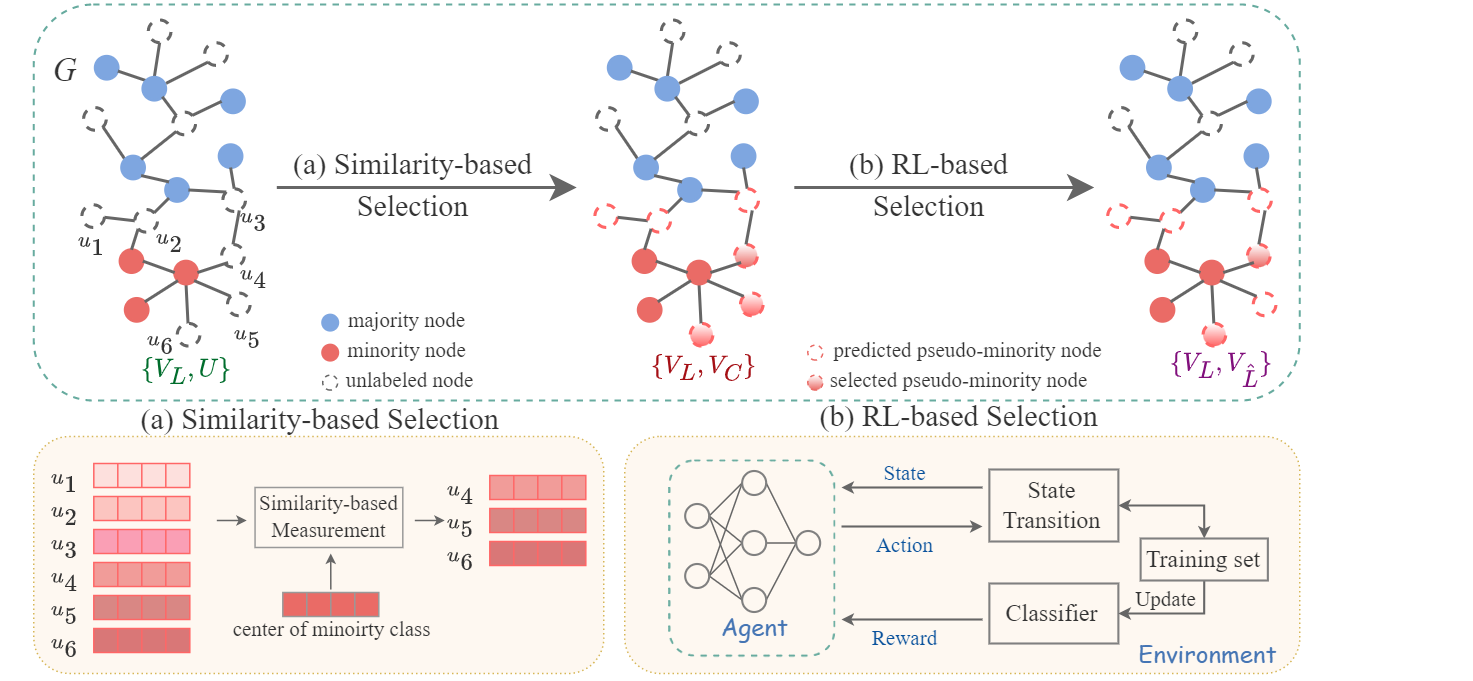}
    \caption{Overall pipeline of GraphSR}
    \label{fig2}
\end{figure*}
In this section, we present the details of the proposed GraphSR, which is based on the self-training technique. 
In fact, self-training \cite{scudder1965probability} is a classical method which is widely used in semi-supervised learning. In principle, the algorithm iteratively trains a model on the available labelled set and uses the trained model to generate pseudo-labels for those unlabelled data; then, it selects confident samples from the unlabelled set to combine with the training set to further retrain the model, until converging.

To accommodate the class-imbalanced issue in graphs, with the idea of self-training, we propose two kinds of components to adaptively select informative and reliable nodes from unlabelled data to supplement the minority classes, as demonstrated in Fig.\ref{fig2}.
First, GraphSR trains a GNN model based on the labelled set $V_L$ and generates pseudo-labels for unlabelled nodes in $U$. 
Then a similarity-based selection module is designed to identify the unlabelled nodes which are most similar to the minority nodes, to filter out candidate nodes set $V_C$ for minority classes.
Next, GraphSR utilizes a module of reinforcement learning to adaptively choose the informative and reliable nodes to get a proper  supplemental set $V_{\hat{L}}$, which can optimally and effectively enrich the diversity of minority classes and finally augment the training set.
With the augmented training data $\{V_L, V_{\hat{L}}\}$, we can train a class-balanced node classifier.
In the following, we will show the details of each component.

\subsection{Similarity-Based Selection}
One simple way to augment the minority classes in the semi-supervised setting is to find the similar unlabelled nodes from the original graph.
In general, node representations derived from GNNs can reflect the inter-class and intra-class relationship of nodes, that is, nodes of the same class will be closer in the embedding space, while nodes from different classes should be farther away in the latent space.
Therefore, rather than comparing nodes directly using their raw attributes, we train a GNN model on the labelled set to learn the node representations, which can simultaneously capture both feature property and topological information of nodes. 
Specifically, we train a GNNs classifier $g$ on imbalanced training set $\{V_L, Y_L\}$, the message passing and fusing process of which is formulated as:
\begin{equation}
    h_v^k = \sigma  \left( W^k \cdot {\rm CAT} (h_v^{k-1}, {\rm AGG} (\{h_{v'}^{k-1}, \forall v' \in \mathcal{N}(v)\})  ) \right)
    \label{eq_mp}
\end{equation}
where ${\rm AGG( \cdot )}$ denotes the aggregation function that aggregate the information of neighborhood $\mathcal{N}(v)$, ${\rm CAT( \cdot )}$ concatenates the node representations and the neighboring information, $W$ is the learnable weight parameters, and $\sigma$ refers to the nonlinear activation function, $h_v^k$ denotes the learned representation of node $v$ with $k$-hop neighbors and $h_v^0=X[v,:]$. 
And we utilize $z_v$ to indicate the embeddings of node $v$ acquired by classifier $g$.

In addition, the well-trained $g$ is able to generate pseudo-labels $\hat{y}$ for unlabelled node $u$, then we can filter out some  unlabelled nodes that may not belong to the minority classes for efficiency.
To this end, let $M_i= \{u \in U|\hat{y}_u = i \}$ denote the set of all the unlabelled nodes which are predicted as minority class $i$ by the classifier $g$. From $M_i$, we only select those nodes that are close enough to the center of the minority class in the embedding space by means of a similarity-based module.
The center of each minority class in the latent space is computed based on the labelled nodes as:
\begin{equation}
    cen(i) = \frac{1}{|C_i|}\sum_{v \in C_i} z_v  
    \label{eq_center}
\end{equation}
where $C_i$ is the labelled nodes set of class $i$. 
Then, the module only selects the top-$K$ nearest nodes to the center $cen(i)$ from $M_i$ as a candidate set $V_C$,  such that
\begin{equation}
    V_C = \{u \in M_i | \mathcal{D}(z_u,cen(i))<\varphi  \ {\rm and} \ |V_C| \leq K \}
    \label{eq_candidate}
\end{equation}
where $\mathcal{D}(\cdot ,\cdot)$ measures the similarity in the embedding space, and we adopt the Euclidean distance as the measurement $\mathcal{D}(u ,v)=\parallel z_u-z_v \parallel $, $\varphi$ denotes the farthest distant between node in $V_C$ and the $cen(i)$, $\varphi = max(z_u,cen(i)|u \in V_C)$.

Through this selection, we can find the nodes which are most likely to be predicted as minority classes, however, $g$ is not reliable because it is trained with the imbalanced data. To deal with this, GraphSR utilizes another selection module of reinforcement learning to draw out the reliable nodes that can exactly improve the performance of classifier, as well as, determine the over-sampling scale for each minority class adaptively.

\subsection{RL-Based Selection}
The key task of this selection module is to specify a sampling procedure that can adaptively select unlabelled nodes to supplement the minority classes. Due to the lack of supervised information on unlabelled nodes, we adopt reinforcement learning for node selection.
We design an iterative sampling procedure which is formulated as a Markov Decision Process (MDP), $ M = (\mathcal{S} , \mathcal{A} , \mathcal{R}, \mathcal{T}) $.
The procedure to generate a balanced training set can be described by a trajectory $(s_0, a_0,r_0, ..., s_T, a_T, r_T)$, where the initial state $s_0$ only contains the imbalanced labelled set and the last state $s_T$ contains the final supplemented balanced nodes set.
GraphSR tries to learn an optimal policy using the reinforcement learning algorithm to allow the agent to decide preserving or discarding the unlabelled nodes in a partially-observed environment. 
In particular, the agent (i.e. the selector) traverses the candidate minority nodes in $V_C$ sequentially. For each node  $u_t$, the agent takes an action through a policy network represented by $\pi_\theta $ based on the current state, and then the environment assigns a reward based on the action. The agent updates the policy network based on the reward.
After enough interactions between the agent and the environment, the agent can learn an optimal policy to optimally select the unlabelled nodes to supplement the minority classes.
With the RL-based selection, it is easier for the algorithm to be generalized to different datasets without additionally determining the over-sampling scale.
In the following, we discuss the major components of RL-based selection module in detail. 

\subsubsection{State}
We define the state $s_t$ of the environment to encode the information of intermediate training set $V_t$ and the unlabelled node $u_t$ at time step $t$.
In order to feed the $s_t$ into the policy network, we need to fix the dimension of $s_t$ irrelevant to the number of nodes in $V_t$.  Motivated by PULNS \cite{luo2021pulns}, we use the summation of the nodes embeddings in $V_t$ to represent the information of $V_t$, i.e.,  $z_{V_t} = \sum\nolimits_{v \in V_t} z_v$.
In addition, the embedding of $u_t$ is exploited to denote its information.
For time step $t$, state $s_t$ is defined as $s_t = (z_{V_t}, z_{u_t})$. At the beginning, $V_0 = V_L$ is the imbalanced labelled set, $u_0$ is the first node in $V_C$. 

\subsubsection{Action}
The action $a_t$ is to decide whether the current unlabelled node $u_t$ in $V_C$ should be included into the current training set $V_t$ or not at time step $t$. Particularly, $a_t \in \{0, 1\}$, where $a_t = 1$ means that node $u_t$ is selected to supplement the imbalanced training set, while $a_t = 0$ indicates that $u_t$ is not applicable.
Furthermore, the action is generated by a policy function $\pi_\theta$, which takes the state as input and is parameterized by $\theta$. In this work, the policy network represents the probability distribution of action, and is specified as a Multilayer Perceptron (MLP) with nonlinear activation function, i.e.:
\begin{equation}
    a_t = P(a_t|s_t)=\pi_\theta(s_t)=MLP_\theta(s_t)
    \label{eq_action}
\end{equation}

\subsubsection{Transition}
After taking action $a_t$, the state of the environment should change to $s_{t+1}$. In our work, state consists of $V_t$ and $u_t$, after taking $a_t$, 
\begin{equation}
    V_{t+1} = \begin{cases}
        \{V_t \cup u_t\}, & a_t = 1 \\
        V_t             , & a_t = 0
    \end{cases}
    \label{eq_trans}
\end{equation}
and then $s_{t+1} = (z_{V_{t+1}}, z_{u_{t+1}})$. 
The termination of transition will happen when agent has completely traversed the candidate set $V_C$ once.

\subsubsection{Reward}
The reward $r_t$ given by environment is to evaluate the action $a_t$ at state $s_t$.
Without supervised information about unlabelled nodes, it is hard to exactly find the minority nodes and reward it explicitly according to the true labels. 
Here we train a classifier based on $\{V_t \cup u_t\}$, and evaluate its accuracy $acc_t$ on a small balanced validation set.
However, the accuracy is always non-negative, directly applying it as reward may impede the convergence of the agent. 
The idea of reward engineering is to assign a positive reward if adding $u_t$ can boost the performance of classifier, or a negative reward otherwise.
Hence, the reward function is designed as:
\begin{equation}
    r_t = \begin{cases}
        +1, & acc_t \geq b_t \ {\rm and} \ a_t = 1 \\
        -1, & acc_t < b_t \ {\rm and} \ a_t = 1 \\
        +1, & acc_t < b_t \ {\rm and} \ a_t = 0 \\
        -1, & acc_t \geq b_t \ {\rm and} \ a_t = 0
    \end{cases}
    \label{eq_reward}
\end{equation}
where, $b_t$ denotes a baseline reward which is the average of past ten accuracies, i.e., $b_t = mean\{acc_{t-11},...,acc_{t-1}\}$, and $acc_0$ means the accuracy of the initial classifier trained by the labelled nodes set $V_L$.

\subsubsection{Policy Gradient Training}
The goal of the agent is to train an optimal policy network that can maximize the expected reward, and policy gradient based methods are broadly utilized to optimize the policy network. 
In this work, we use Proximal Policy Optimization (PPO) \cite{schulman2017proximal} to update the parameter $\theta$ of policy network. The objective function of PPO is defined as:
\begin{equation}
    L^{CLIP}(\theta) = \mathbb{E}_t [{\rm min}(p_t(\theta)\hat{A}_t, {\rm clip}(p_t(\theta), 1-\epsilon,1+\epsilon)\hat{A}_t)]
    \label{eq_ppo}
\end{equation}
where $p_t(\theta)$ is the probability ratio, $p_t(\theta) = \frac{\pi_\theta(a_t|s_t)}{\pi_{\theta_{old}}(a_t|s_t)}$, which is clipped into the range $[1-\epsilon, 1+\epsilon]$, making a lower bound of the conservative policy  iteration objective \cite{kakade2002approximately} and the agent's exploration more stable. 
$\hat{A}_t$ is the estimated advantage function that involves discount accumulated reward and value function $V^\pi$, and is widely used in policy gradient algorithms. 

\subsection{GNN-Based Classifier}
As discussed before, with the similarity-based selection and RL-based selection, GraphSR can sample the most informative and reliable nodes from the unlabelled data to supplement the minority classes for training. 
In detail, we obtain the final training set $\{V_L, V_{\hat{L}}\}$ with labelled set $\{(v_i, y_i)\}$ and supplemented set with pseudo-label $\{(u_i, \hat{y_i})\}$.
Based on the new training data, we can train an unbiased GNN classifier $f$ according to the message passing process as eq.\ref{eq_mp} and cross-entropy loss function. 
The end-to-end training process of GraphSR is outlined in appendix.

\section{Experiments}
We conduct experiments to evaluate the effectiveness of GraphSR for class-imbalanced node classification over several datasets with different imbalanced ratios. The results are reported in this section. More details about the experiments setting and metric definitions are given in appendix.

\begin{table*}[t]
    \renewcommand{\arraystretch}{1.1}
    \centering
    \resizebox{\linewidth}{!}{
    \begin{tabular}{llccccccccc}
    \toprule
                          &               & \multicolumn{3}{c}{Cora} & \multicolumn{3}{c}{CiteSeer} & \multicolumn{3}{c}{PubMed}         \\ \cline{3-11}
                          & Method        & ACC   & F1   & AUC-ROC   & ACC     & F1    & AUC-ROC    & ACC & F1 & AUC-ROC                 \\ 
                          \cline{2-11}
                          \rule{0pt}{2.5ex}
\multirow{10}{*}{\rotatebox{90}{GCN}} 
                          & Vanilla 
                          & 72.25\tiny{$\pm$0.93} & 71.72\tiny{$\pm$1.17} & 88.36\tiny{$\pm$1.19} 
                          & 50.30\tiny{$\pm$2.13} & 43.86\tiny{$\pm$3.02} & 81.79\tiny{$\pm$0.71} 
                          & 64.20\tiny{$\pm$1.34} & 61.14\tiny{$\pm$2.38} & 80.66\tiny{$\pm$0.89}  
                          \\ \cdashline{2-11} \rule{0pt}{2ex}
                          & Re-Weighting  
                          & 72.43\tiny{$\pm$1.31} & 71.82\tiny{$\pm$1.38} & 89.43\tiny{$\pm$1.29} 
                          & 52.23\tiny{$\pm$2.52} & 46.71\tiny{$\pm$3.05} & 82.37\tiny{$\pm$0.60} 
                          & 63.26\tiny{$\pm$1.38} & 60.02\tiny{$\pm$2.14} & 79.98\tiny{$\pm$1.22}   
                          \\
                          & EN-Weighting  
                          & 73.48\tiny{$\pm$1.82} & 72.98\tiny{$\pm$1.79} & 88.59\tiny{$\pm$1.87} 
                          & 51.63\tiny{$\pm$2.46} & 46.49\tiny{$\pm$2.76} & 82.30\tiny{$\pm$2.90} 
                          & 62.67\tiny{$\pm$1.99} & 58.86\tiny{$\pm$3.17} & 79.67\tiny{$\pm$1.49}   
                          \\
                          & Over-Sampling 
                          & 71.82\tiny{$\pm$1.61} & 71.37\tiny{$\pm$1.57} & 87.71\tiny{$\pm$1.28} 
                          & 52.70\tiny{$\pm$1.87} & 47.48\tiny{$\pm$2.38} & 82.39\tiny{$\pm$0.88} 
                          & 63.13\tiny{$\pm$1.81} & 59.50\tiny{$\pm$3.05} & 80.58\tiny{$\pm$1.05}   
                          \\
                          & CB-Sampling   
                          & 67.72\tiny{$\pm$0.64} & 67.07\tiny{$\pm$0.71} & 88.49\tiny{$\pm$0.33} 
                          & 52.26\tiny{$\pm$3.69} & 48.45\tiny{$\pm$3.47} & 81.82\tiny{$\pm$0.86} 
                          & 66.33\tiny{$\pm$3.73} & 64.99\tiny{$\pm$5.49} & 82.37\tiny{$\pm$2.27}  
                          \\
                          & GraphSMOTE    
                          & 68.67\tiny{$\pm$3.56} & 67.66\tiny{$\pm$3.73} & 90.71\tiny{$\pm$1.47} 
                          & 46.83\tiny{$\pm$4.61} & 44.20\tiny{$\pm$5.02} & 76.71\tiny{$\pm$2.85} 
                          & 66.32\tiny{$\pm$3.61} & 64.66\tiny{$\pm$4.68} & 82.18\tiny{$\pm$1.06}   
                          \\
                          & GraphENS      
                          & 73.48\tiny{$\pm$0.13} & 72.95\tiny{$\pm$0.18} & 90.70\tiny{$\pm$0.08} 
                          & 55.75\tiny{$\pm$0.32} & 52.87\tiny{$\pm$0.22} & 82.98\tiny{$\pm$0.06} 
                          & 70.07\tiny{$\pm$0.07} & 69.09\tiny{$\pm$0.34} & 83.45\tiny{$\pm$0.54}   
                          \\ \cline{2-11}
                          & RU-Selection  
                          & 70.75\tiny{$\pm$1.28} & 70.36\tiny{$\pm$1.36} & 89.77\tiny{$\pm$0.71} 
                          & 56.66\tiny{$\pm$0.90} & 53.51\tiny{$\pm$1.03} & 83.21\tiny{$\pm$0.78} 
                          & 69.83\tiny{$\pm$1.36} & 69.32\tiny{$\pm$1.55} & 83.45\tiny{$\pm$0.54}   
                          \\
                          & SU-Selection  
                          & 73.10\tiny{$\pm$0.85} & 72.80\tiny{$\pm$0.91} & \textbf{90.97}\tiny{$\pm$0.22} 
                          & 56.27\tiny{$\pm$1.39} & 53.47\tiny{$\pm$2.24} & 83.49\tiny{$\pm$0.61} 
                          & 67.75\tiny{$\pm$1.01} & 65.83\tiny{$\pm$0.94} & 81.91\tiny{$\pm$0.91}  
                          \\
                          & GraphSR       
                          & \textbf{73.90}\tiny{$\pm$0.16} & \textbf{73.59}\tiny{$\pm$0.11} & 90.21\tiny{$\pm$0.77} 
                          & \textbf{57.28}\tiny{$\pm$0.55} & \textbf{55.20}\tiny{$\pm$0.85} & \textbf{83.67}\tiny{$\pm$0.60} 
                          & \textbf{71.79}\tiny{$\pm$1.34} & \textbf{71.70}\tiny{$\pm$1.41} & \textbf{85.14}\tiny{$\pm$0.86}      
                          \\ 
                          \cline{2-11} 
                          \noalign{\vskip\doublerulesep
                          \vskip-\arrayrulewidth} \cline{2-11}
                          \rule{0pt}{2.5ex} 
    \multirow{10}{*}{\rotatebox{90}{GraphSAGE}} 
                          & Vanilla 
                          & 75.67\tiny{$\pm$0.11} & 75.26\tiny{$\pm$0.11} & 93.71\tiny{$\pm$0.53} 
                          & 49.99\tiny{$\pm$0.52} & 42.75\tiny{$\pm$0.66} & 83.79\tiny{$\pm$0.16} 
                          & 63.67\tiny{$\pm$0.07} & 59.89\tiny{$\pm$0.91} & 86.88\tiny{$\pm$0.16} 
                          \\ \cdashline{2-11}
                          & Re-Weighting  
                          & 76.11\tiny{$\pm$0.46} & 75.99\tiny{$\pm$0.61} & 94.55\tiny{$\pm$0.59} 
                          & 50.99\tiny{$\pm$0.59} & 44.39\tiny{$\pm$0.80} & 77.89\tiny{$\pm$0.42} 
                          & 61.40\tiny{$\pm$0.67} & 57.38\tiny{$\pm$1.17} & 88.41\tiny{$\pm$0.41}
                          \\
                          & EN-Weighting  
                          & 76.05\tiny{$\pm$1.05} & 75.39\tiny{$\pm$0.98} & 94.49\tiny{$\pm$0.24} 
                          & 50.56\tiny{$\pm$0.57} & 43.81\tiny{$\pm$0.93} & 77.98\tiny{$\pm$0.66} 
                          & 61.60\tiny{$\pm$0.71} & 57.25\tiny{$\pm$0.91} & 88.61\tiny{$\pm$1.18} 
                          \\
                          & Over-Sampling 
                          & 75.28\tiny{$\pm$0.56} & 74.86\tiny{$\pm$0.56} & 94.07\tiny{$\pm$0.25} 
                          & 50.66\tiny{$\pm$0.08} & 43.70\tiny{$\pm$0.98} & 84.07\tiny{$\pm$0.31} 
                          & 66.39\tiny{$\pm$1.35} & 64.03\tiny{$\pm$1.71} & 87.07\tiny{$\pm$0.23} 
                          \\
                          & CB-Sampling   
                          & 73.53\tiny{$\pm$0.77} & 72.97\tiny{$\pm$0.90} & 93.64\tiny{$\pm$0.21} 
                          & 52.73\tiny{$\pm$0.75} & 48.67\tiny{$\pm$0.08} & 82.45\tiny{$\pm$0.31} 
                          & 67.13\tiny{$\pm$0.88} & 66.24\tiny{$\pm$0.90} & 87.14\tiny{$\pm$0.20} 
                          \\
                          & GraphSMOTE    
                          & 75.32\tiny{$\pm$0.80} & 75.03\tiny{$\pm$0.87} & 94.42\tiny{$\pm$0.55} 
                          & 43.17\tiny{$\pm$2.87} & 38.85\tiny{$\pm$3.31} & 76.98\tiny{$\pm$1.95} 
                          & 67.22\tiny{$\pm$1.85} & 65.23\tiny{$\pm$2.55} & 85.71\tiny{$\pm$0.29} 
                          \\
                          & GraphENS      
                          & 76.84\tiny{$\pm$0.88} & 75.94\tiny{$\pm$0.84} & 94.04\tiny{$\pm$0.81} 
                          & 52.45\tiny{$\pm$0.23} & 50.93\tiny{$\pm$0.42} & 84.05\tiny{$\pm$0.41} 
                          & 68.07\tiny{$\pm$0.27} & 66.19\tiny{$\pm$0.36} & 88.17\tiny{$\pm$0.22} 
                          \\ \cline{2-11}
                          & RU-Selection  
                          & 75.85\tiny{$\pm$0.82} & 75.61\tiny{$\pm$0.86} & 94.24\tiny{$\pm$0.43} 
                          & \textbf{56.25}\tiny{$\pm$0.2} & \textbf{52.74}\tiny{$\pm$0.49} & \textbf{85.09}\tiny{$\pm$0.26} 
                          & 68.74\tiny{$\pm$0.75} & 68.34\tiny{$\pm$0.79} & 86.72\tiny{$\pm$0.51}   
                          \\
                          & SU-Selection  
                          & 77.99\tiny{$\pm$1.00} & 77.69\tiny{$\pm$1.02} & 94.57\tiny{$\pm$0.25} 
                          & 52.93\tiny{$\pm$0.42} & 47.87\tiny{$\pm$0.87} & 83.79\tiny{$\pm$0.12} 
                          & 66.83\tiny{$\pm$0.37} & 64.72\tiny{$\pm$0.30} & 86.59\tiny{$\pm$0.30}   
                          \\
                          & GraphSR       
                          & \textbf{78.78}\tiny{$\pm$0.42} & \textbf{78.36}\tiny{$\pm$0.45}  & \textbf{94.92}\tiny{$\pm$0.25} 
                          & 54.30\tiny{$\pm$0.52} & 51.15\tiny{$\pm$0.69} & 84.21\tiny{$\pm$0.14} 
                          & \textbf{74.13}\tiny{$\pm$1.19} & \textbf{74.36}\tiny{$\pm$1.34} & \textbf{89.33}\tiny{$\pm$0.38}     
                          \\ 
    \bottomrule
    \end{tabular}}
    \caption{Comparisons of GraphSR with other baselines when imbalance ratio is 0.3. }
    \label{tab:result}
    \end{table*}

\begin{table*}[htbp]
    \renewcommand{\arraystretch}{1.1}
    \centering
    \resizebox{\linewidth}{!}{
    \begin{tabular}{llccc|ccc|ccc|ccc}
    \toprule
                          &  & \multicolumn{3}{c}{$\rho=0.1$} & \multicolumn{3}{c}{$\rho=0.2$} & \multicolumn{3}{c}{$\rho=0.4$} &\multicolumn{3}{c}{$\rho=0.5$}  
                          \\ \cline{3-14}
                          & Method & ACC & F1 & AUC-ROC & ACC & F1 & AUC-ROC & ACC & F1 & AUC-ROC & ACC & F1 & AUC-ROC    
                          \\ \cline{2-14}
                          \rule{0pt}{2.2ex}
                          \multirow{10}{*}{\rotatebox{90}{GraphSAGE}} 
                          & Vanilla 
                          & 61.53\tiny{$\pm$0.53} & 55.40\tiny{$\pm$0.90} & 91.52\tiny{$\pm$1.01} 
                          & 72.67\tiny{$\pm$0.44} & 71.42\tiny{$\pm$0.58} & 93.49\tiny{$\pm$0.55} 
                          & 76.04\tiny{$\pm$0.70} & 75.64\tiny{$\pm$0.77} & 94.69\tiny{$\pm$0.09}
                          & 78.49\tiny{$\pm$0.38} & 78.28\tiny{$\pm$0.39} & 95.09\tiny{$\pm$0.24}   
                          \\ \cdashline{2-14} \rule{0pt}{2ex}
                          & Re-Weighting  
                          & 64.25\tiny{$\pm$0.07} & 58.85\tiny{$\pm$0.78} & 92.72\tiny{$\pm$0.79} 
                          & 74.10\tiny{$\pm$0.61} & 73.37\tiny{$\pm$0.55} & 94.13\tiny{$\pm$0.36} 
                          & 77.13\tiny{$\pm$0.70} & 76.83\tiny{$\pm$0.74} & 95.05\tiny{$\pm$0.09}
                          & 77.85\tiny{$\pm$0.65} & 77.48\tiny{$\pm$0.71} & 95.11\tiny{$\pm$0.21}  
                          \\
                          & EN-Weighting  
                          & 64.04\tiny{$\pm$0.81} & 58.70\tiny{$\pm$0.79} & 93.62\tiny{$\pm$0.25} 
                          & 74.20\tiny{$\pm$0.64} & 73.36\tiny{$\pm$0.73} & 94.31\tiny{$\pm$0.24} 
                          & 76.45\tiny{$\pm$0.38} & 76.02\tiny{$\pm$0.40} & 94.80\tiny{$\pm$0.07}
                          & 77.92\tiny{$\pm$0.83} & 78.09\tiny{$\pm$0.30} & 95.17\tiny{$\pm$0.22}  
                          \\
                          & Over-Sampling 
                          & 64.17\tiny{$\pm$0.76} & 58.55\tiny{$\pm$1.05} & 92.78\tiny{$\pm$0.43} 
                          & 72.81\tiny{$\pm$1.23} & 71.90\tiny{$\pm$1.40} & 93.78\tiny{$\pm$0.18} 
                          & 77.42\tiny{$\pm$0.66} & 77.10\tiny{$\pm$0.67} & 94.81\tiny{$\pm$0.21}
                          & 78.31\tiny{$\pm$0.39} & 78.10\tiny{$\pm$0.38} & 95.11\tiny{$\pm$0.30}  
                          \\
                          & CB-Sampling   
                          & 62.28\tiny{$\pm$0.26} & 56.07\tiny{$\pm$0.19} & 91.50\tiny{$\pm$0.69} 
                          & 72.49\tiny{$\pm$0.59} & 71.40\tiny{$\pm$0.78} & 93.80\tiny{$\pm$0.51} 
                          & 77.10\tiny{$\pm$0.65} & 76.80\tiny{$\pm$0.69} & 95.08\tiny{$\pm$0.17}
                          & 77.03\tiny{$\pm$0.99} & 76.66\tiny{$\pm$1.08} & 95.04\tiny{$\pm$0.31}  
                          \\
                          & GraphSMOTE    
                          & 65.99\tiny{$\pm$2.54} & 63.23\tiny{$\pm$2.70} & 92.35\tiny{$\pm$0.96} 
                          & 71.74\tiny{$\pm$1.92} & 70.61\tiny{$\pm$2.22} & 92.82\tiny{$\pm$1.82} 
                          & 75.54\tiny{$\pm$2.25} & 74.97\tiny{$\pm$2.45} & 94.41\tiny{$\pm$0.92}
                          & 77.72\tiny{$\pm$1.59} & 77.48\tiny{$\pm$1.69} & 94.89\tiny{$\pm$0.42}  
                          \\
                          & GraphENS      
                          & 70.44\tiny{$\pm$0.67} & 65.51\tiny{$\pm$0.88} & 89.55\tiny{$\pm$0.66} 
                          & 75.42\tiny{$\pm$0.11} & 73.74\tiny{$\pm$0.20} & 93.72\tiny{$\pm$0.06} 
                          & 75.84\tiny{$\pm$0.09} & 74.70\tiny{$\pm$0.10} & 94.06\tiny{$\pm$0.01}
                          & 76.26\tiny{$\pm$0.23} & 75.29\tiny{$\pm$0.18} & 94.53\tiny{$\pm$0.18}  
                          \\ \cline{2-14}
                          & RU-Selection  
                          & 71.33\tiny{$\pm$0.91} & 68.37\tiny{$\pm$1.15} & 94.13\tiny{$\pm$0.31} 
                          & 74.74\tiny{$\pm$0.89} & 74.32\tiny{$\pm$0.83} & 93.88\tiny{$\pm$0.58} 
                          & 77.80\tiny{$\pm$0.68} & 77.49\tiny{$\pm$0.62} & 94.70\tiny{$\pm$0.13}
                          & 77.78\tiny{$\pm$0.29} & 77.54\tiny{$\pm$0.22} & 94.96\tiny{$\pm$0.26}  
                          \\
                          & SU-Selection  
                          & 71.39\tiny{$\pm$0.92} & 69.00\tiny{$\pm$1.26} & 94.10\tiny{$\pm$0.50} 
                          & 77.13\tiny{$\pm$0.94} & 76.91\tiny{$\pm$0.96} & 94.70\tiny{$\pm$0.29} 
                          & 77.56\tiny{$\pm$0.23} & 77.27\tiny{$\pm$0.34} & 95.01\tiny{$\pm$0.06}
                          & 78.21\tiny{$\pm$0.41} & 77.96\tiny{$\pm$0.39} & 95.13\tiny{$\pm$0.14}  
                          \\
                          & GraphSR       
                          & \textbf{75.17}\tiny{$\pm$0.85} & \textbf{74.82}\tiny{$\pm$0.86} & \textbf{94.20}\tiny{$\pm$0.12} 
                          & \textbf{77.46}\tiny{$\pm$0.75} & \textbf{77.23}\tiny{$\pm$0.80} & \textbf{94.85}\tiny{$\pm$0.32} 
                          & \textbf{78.56}\tiny{$\pm$0.89} & \textbf{78.25}\tiny{$\pm$0.94} & \textbf{95.33}\tiny{$\pm$0.34}
                          & \textbf{79.54}\tiny{$\pm$0.35} & \textbf{79.33}\tiny{$\pm$0.38} & \textbf{95.50}\tiny{$\pm$0.29}  
                          \\ 
                          
    \bottomrule
    \end{tabular}}
    \caption{Comparisons of GraphSR with other baselines on Cora with different imbalance ratio $\rho$. }
    \label{tab:imratio}
    \end{table*}

\subsection{Experiment Setup}
\subsubsection{Datasets}
We evaluate GraphSR on several widely-used public datasets for node classification task: Cora, CiteSeer, PubMed for citation networks \cite{sen2008collective}.
In citation networks, we construct an imitative imbalanced setting: three classes for Cora and CiteSeer and one class for PubMed are randomly selected as minority classes. All majority classes maintain 20 nodes in the training set, and the numbers for minority classes are $20\times\rho$, where $\rho$ is the imbalanced ratio.
When validating and testing, we sample the same numbers of nodes for all classes to make the validation and test set balanced.
The statistics of datasets are summarized in appendix.

\subsubsection{Baselines}
We test our method over two popular architectures, GCN and GraphSAGE. And we compare GraphSR with representative approaches which handle class-imbalanced issue.
Note that for all the over-sampling baselines algorithms, we oversample the minority classes until they have the same number of samples as that of the majority classes.
\begin{itemize}
    \item Re-Weighting: A classic cost-sensitive method \cite{japkowicz2002class}, which modifies loss function inversely proportional to the number of each class.
    \item EN-Weighting: Another variant of re-weighting method, which assigns the weight for each class based on the Effective Number \cite{cui2019class}.
    \item Over-Sampling: A classical re-sampling method, where the minority nodes are repeatedly sampled until the number of each minority class is the same as that of the majority classes.
    \item CB-Sampling: A variant of re-sampling method motivated by \cite{butler1956machine}, which firstly selects a class among all classes, and then randomly samples a node from the selected class.
    \item GraphSMOTE \cite{zhao2021graphsmote}: An over-sampling method for graph, which synthesizes additional minority nodes from existing nodes in the minority class.
    \item GraphENS \cite{park2021graphens}: Another over-sampling strategy for graph, which generates minority nodes by mixing minority nodes with some nodes sampled from other classes.
    \item RU-Selection: A baseline model that supplements the minority class by randomly collecting the unlabelled nodes whose pseudo-labels are the minority class, until the class distribution is balanced.
    \item SU-Selection: An extension of RU-Selection that, rather than random collection, selects the unlabelled nodes in terms of their similarity for the minority classes.
\end{itemize}

\subsection{Experiment Results}

\subsubsection{Class-Imbalanced Node Classification}
We compare the class-imbalanced node classification performance of GraphSR with that of other baselines on the widely-used citation datasets in semi-supervised setting. 
To verify the model generalization, we combine the proposed technique with two  popular GNNs architectures, GCN and GraghSAGE. 
Here, we set the imbalanced ratio $\rho$ as 0.3, and the experimental results are reported in Table \ref{tab:result}, where we can find that GraphSR outperforms the previous baselines including the re-weighting and re-sampling algorithms.
Our method can effectively select proper unlabelled nodes to supplement the minority classes, which can enrich the diversity of minority classes and avoid overfitting.

\subsubsection{Ablation Study}
We verify two components of GraphSR: similarity-based selection module and RL-based selection module. To this end, we introduce baselines: RU-Selection which supplements minority classes with randomly sampled unlabelled nodes (i.e., without Similarity/without RL), and SU-Selection which utilizes the similarity-based selection (i.e., with Similarity/without RL). 
From Table \ref{tab:result} and \ref{tab:imratio}, we find that RU-Selection can not always work well because it is highly dependent on the predictions of GNN classifier when selecting unlabelled nodes, which is more prone to noisy nodes. 
On the other hand, SU-Selection can further improve the performances with the similarity-based selection module.
With the further learning by RL-based module, GraphSR can identify the most informative and reliable unlabelled nodes for minority augmentation, thus,significantly boost the performances.

\begin{figure}[htbp]
	\captionsetup[subfigure]{justification=centering}
	\centering
	\begin{subfigure}{0.23\textwidth}
		\includegraphics[width=\textwidth]{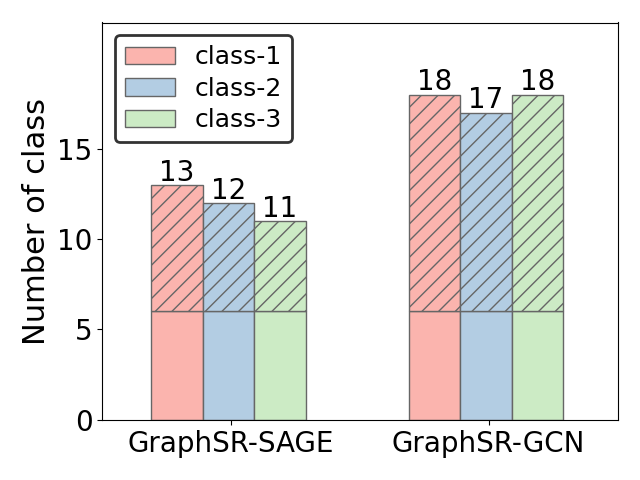}
		\caption{Cora}
		\label{Fig.sup.1}
	\end{subfigure}
	\begin{subfigure}{0.23\textwidth}
		\includegraphics[width=\textwidth]{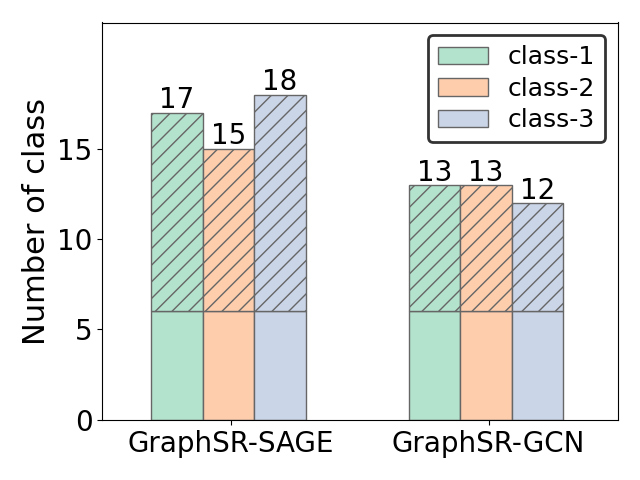}
		\caption{CiteSeer}
		\label{Fig.sup.2}
	\end{subfigure}
	\caption{Over-sampling scale of minority classes with different base architectures on different datasets when imbalanced ratio $\rho=0.3$. Note that the numbers of labelled nodes for minority classes are 6, and the shaded bar indicate the number of unlabelled nodes adaptively determined by GraphSR for each minority class.}
    \label{fig3}
\end{figure}

\subsubsection{Over-Sampling Scale}
Furthermore, we investigate the over-sampling scales of minority classes which are adaptively determined by GraphSR.
We visualize the number of supplemental nodes for each minority class when the imbalance ratio is fixed as 0.3 on Cora and CiteSeer datasets, based on two base architectures. The results are presented in Figure \ref{fig3}, where we can find that GraphSR can automatically determine different over-sampling scales for different minority classes over different datasets.
Besides, even over the same dataset, GraphSR can generate different numbers of supplemental nodes when working for different base architectures, because the over-sampling scales are learned by RL-based selection based on the embeddings of nodes and current training set. 
For different base architectures, the embeddings of nodes are different, which make the over-sampling scale different as well.
With the RL-based selection module, GraphSR can automatically decide the over-sampling scale for different imbalance ratio on different datasets, without demanding a well-elaborating over-sampling hyperparameter.

\subsubsection{Node Classification over Different Imbalance Ratios}
In this subsection, we evaluate the robustness of GraphSR on different imbalance ratio.
The experiments are conducted on Cora dataset with imbalance ratio $\rho$ ranging from 0.1 to 0.5. The results are shown in Table \ref{tab:imratio}, where we can observe that GraphSR can be well adapted to different imbalance ratios. It achieves significant improvements across all different ratios. That demonstrates the validity and robustness of the proposed model.
Moreover, the improvement of GraphSR is more impressive when the degree of imbalanced is extreme. 
On the one hand, the previous over-sampling methods can easily cause overfitting when the minority class is really limited.
On the other hand, the self-training strategy of GraphSR can continuously train the GNN classifer with supplemental training set, and can generate more reliable pseudo-labels for minority classes, which can help GraphSR dig out the more valuable unlabelled nodes.

\section{Conclusion}
In this paper, we investigate the node classification with class-imbalanced problem in a semi-supervised setting.
To take full advantage of rich information from the massive unlabelled nodes, we propose a novel data augmentation strategy, \textit{GraphSR}, which can automatically supplement the minority classes from massive unlabeled nodes, with the help of a similarity-based selection module and an RL-based selection module. 
In addition, the RL-based module can adaptively determine the over-sampling scales for different minority classes.
We verify that the proposed model can effectively enrich the diversity of minority classes and avoid overfitting to some extent.
The experimental results demonstrate the effectiveness and robustness of GraphSR over various datasets with different GNN architectures.

\bibliography{ref}

\clearpage

\subsection{Appendix}

\subsection{Precision and recall on node classification}
\begin{figure*}[htbp]
	\captionsetup[subfigure]{justification=centering}
	\centering
	\begin{subfigure}{0.4\textwidth}
		\includegraphics[width=\textwidth]{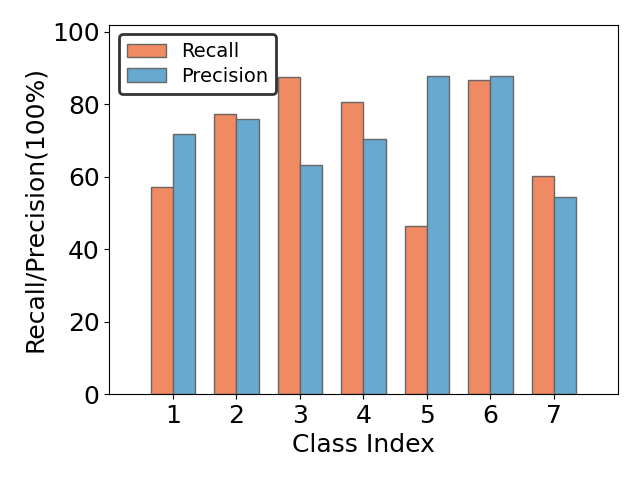}
		\caption{GCN}
		\label{Fig.pr.1}
	\end{subfigure}
	\begin{subfigure}{0.4\textwidth}
		\includegraphics[width=\textwidth]{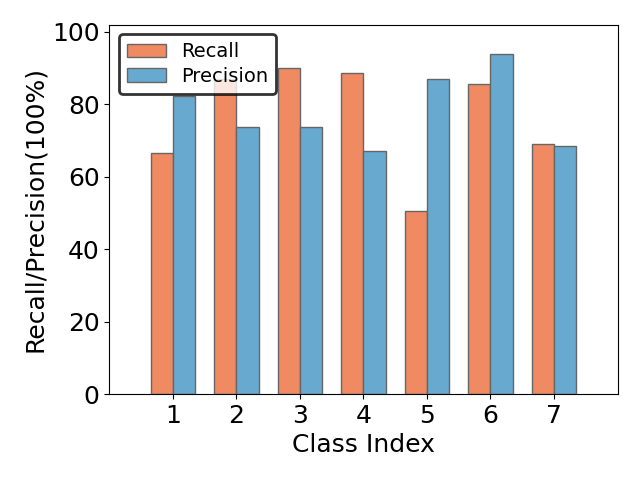}
		\caption{GraphSAGE}
		\label{Fig.pr.2}
	\end{subfigure}
	\caption{Experimental results on Cora dataset. Both GNN models are trained on class-imbalanced data. The test set remains balanced. Note that class $\{1, 2, 3, 4\}$ belong to majority classes, class $\{5, 6, 7\}$ belong to minority classes, and the imbalanced ratio is $\rho=0.3$.
    }
    \label{fig_pr}
    \vspace{-3mm}
\end{figure*}

Here we show the experimental results on Cora dataset to reveal the recall and precision of each class with different GNNs algorithms (GCN and GraphSAGE). We directly apply both algorithms to class-imbalanced semi-supervised setting on Cora dataset, and the results are displayed in Figure \ref{fig_pr}. 
In Cora dataset, we set the first four classes belong to majority classes and the last three classes are minority classes.
CReST \cite{wei2021crest} experimentally finds the phenomenon that minority classes suffer from low recall but achieve surprisingly high precision, thus minority pseudo-labels are less risky to supplement the training set. 
However, we find that this experimental phenomenon is not suitable for all datasets, especially those from different domains.
Specifically, the recalls of majority classes are not always larger than minority classes, for example, the recall of class 6 in Fig. \ref{Fig.pr.1} is higher than majority classes.
Besides, Fig. \ref{Fig.pr.1} and Fig. \ref{Fig.pr.2} show the precision of class 1 (from the majority classes) is even higher than the recall, while the recall of class 7 (from the minority classes) is higher than its precision, which are both contrary to the phenomenon observed by CReST. 
These results indicate that when training classifier with imbalanced data, the predictions on minority classes are not reliable enough, and noisy nodes are easily obtained to supplement the minority classes.
Therefore, it is necessary to carefully determine the supplemental unlabelled nodes.

\subsection{Datasets.}
Here we show the details of datasets in Table \ref{tab:dataset}, Cora\footnote{https://linqs-data.soe.ucsc.edu/public/lbc/cora.tgz}, CiteSeer\footnote{https://linqs-data.soe.ucsc.edu/public/lbc/citeseer.tgz} and PubMed\footnote{https://linqs-data.soe.ucsc.edu/public/Pubmed-Diabetes.tgz} \cite{sen2008collective}.
For citation networks, one node represents a publication, and each node is characterized as a 0/1-valued word vector indicating the absence/presence of the corresponding word in the dictionary, which consists of distinct unique words.
An edge is a citation link between two nodes.
Each node belongs to a research area, i.e., class. And the label distributions of datasets are displays in Table \ref{tab:distribution}. 
In addition, we show data segmentation for each dataset in Table \ref{tab:datasplit}, respectively. The first row indicates the number of majority classes and minority classes. The second and third rows denote the numbers of each class used for training. And all the models are validated and tested in the class-balanced set.
\begin{table}[htbp]
    \renewcommand{\arraystretch}{1.1}
    \centering
    \resizebox{\linewidth}{!}{
    \begin{tabular}{cccc}
    \toprule
    Dataset     & Cora & CiteSeer & PubMed   \\ 
    \midrule
    Number of Nodes        & 2708   & 3327     & 19717          \\
    Number of Features     & 1433   & 3703     & 500          \\
    Number of Edges        & 5429   & 4732     & 44338        \\
    Number of Classes      & 7      & 6        & 3                \\ 
    \bottomrule
    \end{tabular}}
    \caption{Datesets statistics.}
    \label{tab:dataset}
    \end{table}

\begin{table}[htbp]
    \renewcommand{\arraystretch}{1.1}
    \centering
    \resizebox{\linewidth}{!}{
    \begin{tabular}{cccccccc}
    \toprule
    Dataset & \#$L_0$ & \#$L_1$ & \#$L_2$ & \#$L_3$ & \#$L_4$ & \#$L_5$ & \#$L_6$ \\
    \midrule
    Cora     & 818  & 180  & 217  & 426  & 351 & 418 & 298     \\
    Citeseer & 596  & 668  & 701  & 264  & 508 & 590 & -       \\
    Pubmed   & 4103 & 7875 & 7739 & -    & -   & -   & -       \\
    \bottomrule
    \end{tabular}}
    \caption{Label distributions of dataset.}
    \label{tab:distribution}
    \end{table}

\begin{table}[htbp]
    \renewcommand{\arraystretch}{1.1}
    \centering
    \resizebox{\linewidth}{!}{
    \begin{tabular}{cccc}
    \toprule
    Dataset     & Cora & CiteSeer & PubMed   \\ 
    \midrule
    Class Split[Maj/Min]      & 4 / 3      & 3 / 3        & 2 / 1     \\ 
    Number of Majority Class  & 20      & 20        & 20  \\ 
    Number of Minority Class  & $20\times\rho$  & $20\times\rho$   & $20\times\rho$  \\  
    Number of Validation Per Class & 30      & 30        & 30  \\
    Number of Test Per Class & 100      & 100        & 100  \\
    \bottomrule
    \end{tabular}}
    \caption{Date segmentation.}
    \label{tab:datasplit}
    \end{table}

\subsection{Evaluation Metrics}
Following previous works on class-imbalanced classification, such as GraphSMOTE \cite{zhao2021graphsmote}, we adopt three widely-used metrics to measure the performances of all methods, namely accuracy (ACC), F1-macro score (F1) and AUC-ROC score.
ACC is one of the most commonly used metrics in machine learning to measure the performance of model and is susceptible to data imbalance.
F1 is used to measure the performance of imbalanced data, and is considered as a weighted average of the precision and recall of the model.
AUC-ROC is the area under the ROC curve and indicates the probability that the predicted positive cases rank ahead of the negative cases, it is not sensitive to data imbalance.

\subsection{Experimental Setting}
We adopt a 3-layer MLP as the policy network with 128 dimensions in the hidden layer, activated using ReLu function. We use the PPO algorithm to train the agent with default hyperparameters, and the learning rate is set as 0.005. 
All the parameters are optimized with Adam optimizer \cite{kingma2014adam}.
For GNN classifier, we adopt two base architectures GCN and GraphSAGE. Both consist of 2 GCN or SAGE layers with 128 hidden units and ReLU activation. The learning rate is set as 0.01, and dropout rate is 0.5. GNN classifiers are trained by SGD optimizer with early stop mechanism and the maximum training epoch is 2000.
Specifically, the size of candidate set $K$ in similarity-based selection module is set as 20.
For all the compared methods, we report the mean and standard deviation of 5 runs.

\subsection{Experimental Result}
The experimental results of GraphSR based on GAT architecture are reported in Table \ref{tab:gatresult}.

\begin{table*}[t]
    \renewcommand{\arraystretch}{1.1}
    \centering
    \resizebox{\linewidth}{!}{
    \begin{tabular}{llccccccccc}
    \toprule
                          &               & \multicolumn{3}{c}{Cora} & \multicolumn{3}{c}{CiteSeer} & \multicolumn{3}{c}{PubMed}         \\ \cline{3-11}
                          & Method        & ACC   & F1   & AUC-ROC   & ACC     & F1    & AUC-ROC    & ACC & F1 & AUC-ROC                 \\ 
                          \cline{2-11}
                          \rule{0pt}{2.5ex}
\multirow{10}{*}{\rotatebox{90}{GAT}} 
                          & Vanilla 
                          & 75.03\tiny{$\pm$0.94} & 73.67\tiny{$\pm$1.05} & 94.76\tiny{$\pm$0.37} 
                          & 47.49\tiny{$\pm$1.21} & 39.90\tiny{$\pm$2.17} & 81.59\tiny{$\pm$0.61} 
                          & 61.50\tiny{$\pm$0.89} & 57.42\tiny{$\pm$1.14} & 86.08\tiny{$\pm$0.18}  
                          \\ \cdashline{2-11} \rule{0pt}{2ex}
                          & Re-Weighting  
                          & 75.07\tiny{$\pm$0.97} & 74.82\tiny{$\pm$1.11} & 95.45\tiny{$\pm$0.15} 
                          & 50.50\tiny{$\pm$2.01} & 44.96\tiny{$\pm$2.86} & 83.09\tiny{$\pm$0.45} 
                          & 66.83\tiny{$\pm$0.72} & 65.87\tiny{$\pm$0.87} & 86.94\tiny{$\pm$0.11}   
                          \\
                          & EN-Weighting  
                          & 75.91\tiny{$\pm$1.38} & 75.60\tiny{$\pm$1.24} & 95.66\tiny{$\pm$0.15} 
                          & 52.72\tiny{$\pm$0.41} & 47.52\tiny{$\pm$0.36} & 82.90\tiny{$\pm$0.55} 
                          & 65.77\tiny{$\pm$0.56} & 64.44\tiny{$\pm$0.46} & 86.78\tiny{$\pm$0.14}   
                          \\
                          & Over-Sampling 
                          & 75.71\tiny{$\pm$0.76} & 75.37\tiny{$\pm$0.84} & 95.59\tiny{$\pm$0.17} 
                          & 51.54\tiny{$\pm$1.02} & 45.62\tiny{$\pm$1.51} & 83.09\tiny{$\pm$0.48} 
                          & 64.83\tiny{$\pm$1.25} & 62.95\tiny{$\pm$1.69} & 86.55\tiny{$\pm$0.26}   
                          \\
                          & CB-Sampling   
                          & 75.51\tiny{$\pm$0.49} & 75.14\tiny{$\pm$0.52} & 95.39\tiny{$\pm$0.21} 
                          & 49.46\tiny{$\pm$1.54} & 43.01\tiny{$\pm$2.09} & 82.36\tiny{$\pm$0.58} 
                          & 64.66\tiny{$\pm$0.81} & 63.07\tiny{$\pm$0.98} & 86.92\tiny{$\pm$0.24}  
                          \\
                          & GraphSMOTE    
                          & 67.25\tiny{$\pm$3.37} & 66.47\tiny{$\pm$3.27} & 90.17\tiny{$\pm$1.58} 
                          & 54.28\tiny{$\pm$1.69} & 53.16\tiny{$\pm$2.20} & 80.51\tiny{$\pm$0.08} 
                          & 71.11\tiny{$\pm$2.88} & 70.91\tiny{$\pm$3.15} & 86.92\tiny{$\pm$0.57}   
                          \\
                          & GraphENS      
                          & 77.62\tiny{$\pm$1.01} & 76.93\tiny{$\pm$1.02} & 93.23\tiny{$\pm$0.35} 
                          & 57.20\tiny{$\pm$0.98} & 55.03\tiny{$\pm$1.19} & 83.02\tiny{$\pm$0.71} 
                          & 74.90\tiny{$\pm$0.83} & 74.64\tiny{$\pm$0.88} & 88.65\tiny{$\pm$0.55}   
                          \\ \cline{2-11}
                          & RU-Selection  
                          & 77.50\tiny{$\pm$0.42} & 77.23\tiny{$\pm$0.43} & 96.26\tiny{$\pm$0.19} 
                          & 51.12\tiny{$\pm$1.41} & 45.01\tiny{$\pm$1.50} & 83.08\tiny{$\pm$0.99} 
                          & 68.50\tiny{$\pm$0.59} & 67.78\tiny{$\pm$0.53} & 87.69\tiny{$\pm$0.34}   
                          \\
                          & SU-Selection  
                          & 77.31\tiny{$\pm$0.30} & 77.15\tiny{$\pm$0.26} & \textbf{96.31}\tiny{$\pm$0.20} 
                          & 52.96\tiny{$\pm$2.59} & 45.83\tiny{$\pm$1.45} & 82.93\tiny{$\pm$0.86} 
                          & 69.25\tiny{$\pm$0.82} & 69.01\tiny{$\pm$0.86} & 88.29\tiny{$\pm$0.22}  
                          \\
                          & GraphSR       
                          & \textbf{79.64}\tiny{$\pm$0.78} & \textbf{79.44}\tiny{$\pm$0.88} & 95.68\tiny{$\pm$0.19} 
                          & \textbf{57.77}\tiny{$\pm$0.20} & \textbf{55.31}\tiny{$\pm$0.21} & \textbf{84.12}\tiny{$\pm$0.31} 
                          & \textbf{77.88}\tiny{$\pm$0.63} & \textbf{77.82}\tiny{$\pm$0.72} & \textbf{91.68}\tiny{$\pm$0.51}      
                          \\ 

    \bottomrule
    \end{tabular}}
    \caption{Comparisons of GraphSR with other baselines when imbalance ratio is 0.3. }
    \label{tab:gatresult}
    \end{table*}

\subsection{Main Algorithm}
The end-to-end training process of GraphSR is outlined in Algorithm \ref{alg:algorithm}.

\begin{algorithm*}[t]
    \caption{The overall process of GraphSR}
    \label{alg:algorithm}
    \textbf{Input}: $\mathcal{G}=(V,E,X)$; Imbalanced training set with labels $\{V_L, Y_L\}$; Unlabelled node set $U$; Number of epochs $N_e$; Label set of minority class $Y_M$\\
    \textbf{Output}: Augmented training set, unbiased GNN classifier $f$
    \begin{algorithmic}[1] 
    \STATE Initialize a GNN classifier $g$ trained with $V_L$
    \STATE Obtain embeddings of labelled and unlabelled nodes by $g$

    \FOR {$i \in Y_M$}
    \STATE Calculate the center $cen_i$ for  class $i$
    \STATE Obtain pseudo-minority node set $M_i$
    \FOR {$j =1, ... , |M_i|$}
    \STATE Update candidate node set $V_C$ 
    \ENDFOR
    \ENDFOR
    
    \STATE Initialize agent $\pi_\theta$ randomly
    \STATE Initialize current training set $V_t = V_L$
    \FOR{$e=1,...,N_e$}
    \FOR{$t=1,...,|V_C|$}
    \STATE Get state $s_t$
    \STATE Sample action $a_t \sim \pi_\theta(s_t)$
    \STATE Update $V_t$
    \STATE Obtain reward $r_t$
    \ENDFOR
    \STATE Update $\pi_\theta$
    \ENDFOR
    
    \STATE Obtain final augmented training set $\{V_L, V_{\hat{L}}\}$
    \STATE Train an unbiased GNN classifier $f$ based on $\{V_L, V_{\hat{L}}\}$
    \STATE \textbf{return} Agent $\pi_\theta$, GNN classifier $f$
    \end{algorithmic}
    \end{algorithm*}

\end{document}